%% file: main.tex
\documentclass[10pt,twocolumn,letterpaper]{article}

\usepackage[applications]{wacv}


\definecolor{wacvblue}{rgb}{0.21,0.49,0.74}
\usepackage[pagebackref,breaklinks,colorlinks,allcolors=wacvblue]{hyperref}
\def\wacvPaperID{*****} 
\def\confName{WACV}
\def\confYear{2027}

\title{Forehead ROI Localization under Real-World Imaging Degradations via Cross-View Pseudo-Labeling}
\author{
Geetanjali Sharma$^{1,3}$ \quad
Gaurav Jaswal$^{2}$ \quad
Aditya Nigam$^{1}$ \quad
Raghavendra Ramachandra$^{3}$\\
\\
$^{1}$Indian Institute of Technology Mandi, India\\
$^{2}$Division of Digital Forensics, Directorate of Forensic Services Shimla, India \\
$^{3}$SAFE Center, Norwegian University of Science and Technology (NTNU), Gjøvik, Norway
}

\begin{document}
\maketitle
\input{sec/0_abstract}  
\input{sec/1_intro}
\input{sec/2_related_work}
\input{sec/3_framework}
\input{sec/4_experiments}

{
    \small
    \bibliographystyle{ieeenat_fullname}
    \bibliography{main}
}

\end{document}

%% file: sec/0_abstract.tex
\begin{abstract}

Forehead region-of-interest (ROI) localization is important for forehead-creases biometrics, particularly when the lower face is occluded. However, localization remains challenging under varying image quality, pose, and illumination. We propose Confidence and Consistency Guided Pseudo-Labeling (CCGP), which adapts Grounding DINO using unlabeled facial images filtered by detection confidence and geometric consistency between original and horizontally flipped views, without modifying the detector architecture or loss. We evaluate CCGP on five external datasets and 350 images spanning seven imaging conditions. CCGP achieves the highest mean intersection over union (mIoU) on four of five external datasets, improving from 0.415 to 0.580 on RMFRD compared with supervised Grounding DINO. On the quality-stratified subset, the largest gains occur under blur (0.487 to 0.719) and low contrast (0.615 to 0.710). CCGP also outperforms conventional self-training on RMFRD (0.580 versus 0.551 mIoU). Despite a slight in-domain performance gap (0.861 vs. 0.866 mIoU) and challenges with off-angle poses, CCGP improves cross-dataset localization under challenging imaging conditions.
\end{abstract}

%% file: sec/1_intro.tex
\section{Introduction}
\label{sec:intro}

Reliable forehead localization is important for facial analysis applications\cite{schroff2015facenet}, including forehead-creases biometric recognition\cite{bharadwaj2022mobile} \cite{10593960} \cite{tandon2025generating} \cite{tandon2024synthetic} \cite{tandon2025creasegen} \cite{sharmaseeing} \cite{tandon2025fc} and surveillance. The forehead can remain visible when other facial regions, particularly the lower face, are occluded by masks. However, localizing the forehead in unconstrained images remains challenging due to variations in illumination, viewpoint, image quality, and partial occlusion. These variations can reduce localization accuracy and limit the reliability of subsequent forehead-based analysis.
Recent vision-language models, such as Grounding DINO, enable open-set object localization by combining visual features with textual prompts\cite{liu2024eccv-grounding}. This capability provides a promising starting point for forehead localization without requiring a detector designed exclusively for this region. Nevertheless, adapting a
pretrained detector to a specialized localization task requires reliable annotations. Manually annotating large-scale facial datasets is expensive, while directly using automatically generated pseudo labels \cite{li2023robustteacher} \cite{shehzadi2026ssod} can introduce localization errors. Such errors may propagate during fine-tuning and negatively
affect the resulting detector. Therefore, effectively exploiting unlabeled facial images requires a strategy that identifies reliable pseudo-labels while filtering potentially inaccurate predictions.
This work investigates two questions: (i) how does forehead ROI
localization performance vary across real-world imaging conditions
and datasets, and (ii) can cross-view consistent pseudo-labeling
improve localization when adapting a pretrained detector using
unlabeled facial images? We address these questions through
cross-dataset evaluation, quality stratified robustness experiments,
and a comparison with conventional self-training.

In this work, we propose \textbf{Confidence and Consistency Guided Pseudo-Labeling (CCGP)}, a framework for improving forehead localization by leveraging unlabeled facial images. CCGP generates predictions from the original image and its horizontally flipped counterpart, maps both predictions to a common coordinate system, and assesses
their geometric consistency alongside detection confidence. Pseudo-labels satisfying predefined confidence and consistency
criteria are retained, while unreliable predictions are discarded. The accepted pseudo-labels are then combined with
ground-truth annotations from FH-V1 to fine-tune Grounding DINO. Importantly, CCGP does not introduce a new detector architecture or an additional training loss; it improves the fine-tuning data through multi-view pseudo-label selection.
We evaluate CCGP against supervised and conventional self-training baselines to examine the effectiveness of
confidence and consistency guided pseudo-label selection. We further investigate localization performance under challenging imaging conditions and evaluate generalization across datasets. These experiments assess whether filtering pseudo-labels using confidence and cross-view geometric consistency can improve the use of unlabeled data for forehead localization.
The main contributions of this work are summarized as follows:
\begin{itemize}
    \item We propose \textbf{CCGP}, which selects reliable forehead pseudo-labels using detection confidence and cross-view geometric consistency.

    \item We fine-tune Grounding DINO using selected pseudo-labels and manually annotated data, reducing the need for additional manual annotations.

    \item We evaluate CCGP against supervised and self-training baselines in terms of cross-dataset generalization, robustness, and computational efficiency.
\end{itemize}

%% file: sec/2_related_work.tex
\section{Related Work}
\label{sec:related_work}

\subsection{Facial Region Analysis under Occlusion}
\label{subsec:facial_region_analysis}

Facial analysis in real-world scenarios is challenged by occlusion, head pose, illumination, facial scale, and image quality. The Real-World Masked Face Recognition Dataset (RMFRD) \cite{wang2023masked} supports research on face recognition under mask-induced occlusion. However, masked-face recognition identifies individuals, whereas forehead ROI localization determines the spatial extent of the forehead.
Facial region parsing assigns pixel-level semantic labels to facial components. The LaPa dataset and boundary attention based segmentation methods support research on facial parsing and boundary delineation \cite{liu2020lapa}. However, general face parsing differs from forehead ROI localization, which must identify the forehead while distinguishing it from adjacent regions, including the eyebrows, temples, and hairline. These challenges motivate evaluating forehead localization under variations in pose, scale, illumination, and image quality. 

\subsection{Object Detection and Instance Segmentation}
\label{subsec:detection_segmentation}

Object detection\cite{wang2023yolov7}\cite{jocher2023ultralytics} and instance segmentation\cite{gu2024diffusioninst} provide complementary approaches to localizing visual regions. Object detectors predict bounding boxes that specify the approximate spatial extent of a target, whereas instance segmentation methods additionally predict pixel level masks. Bounding box detectors localize the forehead region, whereas segmentation methods predict pixel-level boundaries. Our method performs bounding-box localization rather than pixel-level segmentation.
YOLO's detectors\cite{wang2023yolov7}\cite{jocher2023ultralytics} are attractive for applications requiring efficient inference. YOLOv7\cite{wang2023yolov7} combines architectural and training optimizations to improve the speed accuracy trade-off in real time object detection \cite{wang2023yolov7}. YOLOv8\cite{jocher2023ultralytics} provides a more recent implementation supporting object detection and instance segmentation \cite{jocher2023ultralytics}. These architectures provide useful baselines for investigating the localization accuracy and computational efficiency achievable with single stage approaches. 
Mask R-CNN~\cite{he2017mask} extends Fast R-CNN~\cite{girshick2015fast} with a parallel mask prediction branch, enabling joint instance localization and pixel level segmentation. However, it is not specifically designed for anatomical facial regions, where forehead boundaries can be affected by hairline visibility, eyebrow proximity, head pose, and image quality. Its performance on forehead-specific annotations must therefore be evaluated.

\subsection{Open-Vocabulary Detection and Diffusion-Based Segmentation}
\label{subsec:open_vocab_diffusion}

Open-vocabulary detection offers an alternative to conventional closed-set detectors by allowing target concepts to be specified through natural language expressions. Grounding DINO\cite{liu2024eccv-grounding} combines Transformer based object detection with grounded vision language pretraining for open-set object detection \cite{liu2024eccv-grounding}. This capability is relevant to forehead localization because a target region can be described using a textual prompt without requiring the detector to be restricted to a fixed set of object categories. 
However, open-vocabulary detectors such as Grounding DINO~\cite{liu2024eccv-grounding} predict bounding boxes rather than pixel-level masks. Any masks derived from these boxes require separate evaluation, particularly when compared with instance segmentation models. DiffusionInst~\cite{gu2024diffusioninst} uses a denoising process to predict instance masks. Their suitability for forehead ROI extraction remains underexplored, motivating comparisons of localization accuracy, mask quality, and computational cost. Semi-supervised detection uses confidence-based pseudo-labeling and cross-view consistency. We adapt these strategies to forehead ROI localization using an open-vocabulary detector.

\subsection{Research Gap and Motivation}
\label{subsec:research_gap}

Despite advances in face parsing, object detection, instance segmentation, open-vocabulary localization, and diffusion-based segmentation, dedicated forehead ROI localization remains underexplored. Existing methods target general facial regions or object categories, and their effectiveness under diverse real-world imaging conditions requires systematic evaluation.
Forehead localization is challenged by variations in pose, scale, illumination, shadows, blur, low contrast, occlusion, and partial forehead visibility. Moreover, Bounding box localization and pixel-level segmentation differ; we evaluate localization using box derived overlap metrics, robustness, and computational efficiency.
To investigate these challenges, we evaluate YOLO-based models~\cite{wang2023yolov7,jocher2023ultralytics}, Mask R-CNN~\cite{he2017mask}, Grounding DINO\cite{liu2024eccv-grounding}, and DiffusionInst~\cite{gu2024diffusioninst}. Cross-dataset evaluation uses FH-V2~\cite{tandon2024synthetic}, AgeDB~\cite{moschoglou2017agedb}, LFW~\cite{dalkran2021lfwmasked}, CMFD~\cite{cabani2021maskedface}, and RMFRD~\cite{wang2023masked}. Robustness is assessed on a balanced subset of 350 images covering seven conditions: blur, scale variation, illumination variation, shadows, low contrast, off-angle pose, and good quality images. We further analyze computational efficiency to characterize the trade-offs between accuracy, robustness, and inference cost. 

%% file: sec/3_framework.tex
\section{Proposed Method}
\label{sec:proposed_method}

We propose Confidence and Consistency Guided Pseudo-Labeling (CCGP), a
self-training framework for forehead localization that
exploits cross-view prediction consistency to select
reliable pseudo-labels from unlabeled facial images.
The central motivation is that confidence based
pseudo-label selection alone may retain predictions
that are locally confident but geometrically unstable
under image transformations. CCGP addresses this
limitation by jointly considering detection confidence
and geometric agreement across multiple views before
incorporating automatically generated annotations
into the training set.

\begin{figure}[t]
    \centering
    \includegraphics[width=0.5\textwidth,height=9cm,keepaspectratio]{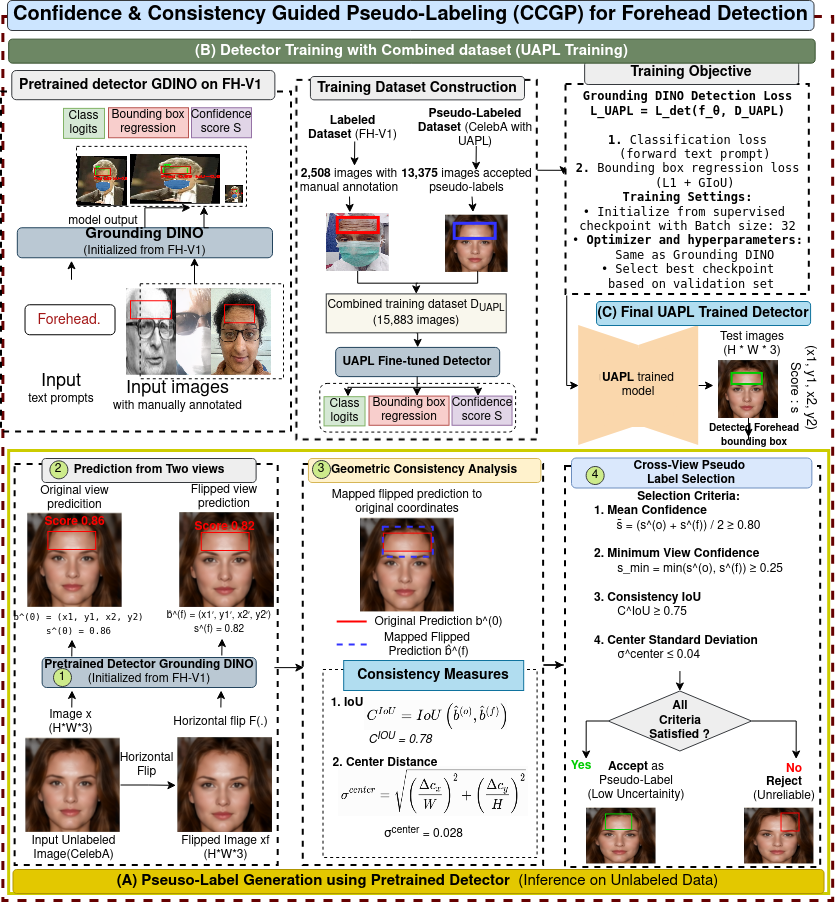}
    \caption{Proposed CCGP framework for confident \& consistency guided pseudo-labeling generation, model fine-tuning, and forehead detection.}
    \label{fig:ccgp_architecture}
\end{figure}

As illustrated in Figure~\ref{fig:ccgp_architecture}, the
proposed framework consists of four stages:
(i) supervised detector initialization,
(ii) multi-view inference,
(iii) confidence and consistency guided pseudo-label selection, and
(iv) detector fine-tuning using a combination of
ground-truth and pseudo-labeled data. Grounding DINO
serves as the underlying detection architecture
throughout the pipeline. The proposed method does not
introduce a separate teacher student architecture or
modify the detector backbone and detection head.
Instead, CCGP improves the pseudo-label generation
procedure by filtering candidate annotations using
confidence and cross-view geometric consistency.

\subsection{Problem Formulation}
\label{sec:problem_formulation}

Let
$\mathcal{D}_{l} =
\{(x_j,b_j^{gt})\}_{j=1}^{N_l}$
denote a labeled forehead dataset, where $x_j$ is an
input image and $b_j^{gt}$ is its ground-truth forehead
bounding box. Let
$\mathcal{D}_{u}=\{x_i\}_{i=1}^{N_u}$
denote an unlabeled dataset.
Our objective is to exploit the unlabeled data to
improve forehead localization while limiting the
introduction of incorrect pseudo-labels. We first
initialize a Grounding DINO detector using the
supervised labeled dataset. The initialized detector
generates candidate forehead annotations for unlabeled
images, which are evaluated using detection confidence
and cross-view geometric consistency. The accepted
pseudo-labels form a set $\mathcal{D}_{p}$, which is
combined with the labeled data to fine-tune the detector.

\subsection{Supervised Detector Initialization}
\label{sec:detector_initialization}

We initialize Grounding DINO using a checkpoint trained on the labeled forehead dataset. Given an image $x$ and the text prompt $q=\texttt{forehead.}$, the detector predicts candidate bounding boxes and their associated confidence scores.
Let $f_{\theta_0}$ denote the initialized detector, where $\theta_0$ represents its pretrained parameters.
We use this detector to generate pseudo-labels for the unlabeled dataset without requiring manual annotation
of every candidate image.
Although supervised initialization provides task specific parameters for forehead localization, a single prediction confidence score does not explicitly measure whether the predicted location remains stable under transformations of the input image. CCGP therefore introduces cross-view consistency analysis to provide an additional criterion for assessing pseudo-label reliability.

\subsection{Multi-View Inference}
\label{sec:multi_view_inference}

For each unlabeled image $x_i$, we construct two views: the original image and its horizontally flipped
counterpart. They are defined as $x_i^{(o)}=x_i, x_i^{(f)}=\mathcal{F}(x_i),$ where $\mathcal{F}(\cdot)$ denotes horizontal flipping.
The initialized detector processes both views using the same text prompt and model parameters. The resulting predictions are $(\hat{b}_i^{(o)},s_i^{(o)}) = f_{\theta_0}(x_i^{(o)},q),$ $(\hat{b}_i^{(f)},s_i^{(f)}) = f_{\theta_0}(x_i^{(f)},q),$
where $\hat{b}_i^{(o)}$ and $\hat{b}_i^{(f)}$ denote the predicted forehead bounding boxes, and $s_i^{(o)}$ and $s_i^{(f)}$ denote their detection confidence scores.
The two views provide complementary predictions of the same forehead region. If a prediction is sensitive
to horizontal reflection, the resulting bounding boxes may disagree after coordinate restoration. CCGP uses
this disagreement as a proxy for localization uncertainty. Horizontal flipping is selected as a simple, label preserving transformation that enables direct geometric consistency checks after restoring the predicted bounding boxes to the original coordinate system. Other transformations, such as rotation and translation, may also improve robustness; however, their effects are not investigated in this study and are left for future work.

\subsection{Geometric Consistency Analysis}
\label{sec:geometric_consistency}

Predictions obtained from the flipped image are expressed in
the flipped-image coordinate system. Therefore, the flipped-view
prediction must first be mapped back to the original coordinate
system. For an image of width $W$, let a bounding box be
$b=(x_1,y_1,x_2,y_2)$. The inverse horizontal-flip transformation
is defined as
$\mathcal{F}^{-1}(b) = (W-x_2,\ y_1,\ W-x_1,\ y_2)$.
The restored flipped-view prediction is consequently
$\tilde{b}_i^{(f)} = \mathcal{F}^{-1}\left(\hat{b}_i^{(f)}\right)$.
We quantify the geometric agreement between the original
and restored flipped-view predictions using intersection
over union (IoU):
\begin{equation}
C_i^{\mathrm{IoU}}
=
\frac{
\left|\hat{b}_i^{(o)}\cap\tilde{b}_i^{(f)}\right|
}{
\left|\hat{b}_i^{(o)}\cup\tilde{b}_i^{(f)}\right|
}.
\label{eq:cross_view_iou}
\end{equation}

A larger value of $C_i^{\mathrm{IoU}}$ indicates stronger
spatial agreement between the original and restored flipped-view
predictions. In addition to box overlap, CCGP employs a
center-disagreement criterion to reject predictions whose
estimated locations vary across views. Let
$(c_{x,i}^{(k)},c_{y,i}^{(k)})$ denote the normalized center
coordinates of the predicted box for view $k$. The center
disagreement is computed as
\begin{equation}
\sigma_i^{\mathrm{center}}
=
\sqrt{
\frac{
\operatorname{Var}_k(c_{x,i}^{(k)})
+
\operatorname{Var}_k(c_{y,i}^{(k)})
}{2}
},
\label{eq:center_std}
\end{equation}
where the variances are computed across the available views.
A pseudo-label satisfies the center-consistency criterion only
if
$\sigma_i^{\mathrm{center}}\leq\tau_{\mathrm{center}}=0.04$.
The threshold of 0.04 was set as a tolerance for normalized
centre displacement between the original and flipped views,
allowing minor localization variations while identifying
substantial positional disagreement.
Cross-view consistency does not guarantee that a prediction
matches the unknown ground truth. Instead, it provides an
additional reliability criterion that complements the
detector's confidence score.
\subsection{Confidence \& Consistency guided  Pseudo-Label Selection}
\label{sec:pseudo_label_selection}

For each unlabeled image, CCGP computes the mean confidence across
the original and horizontally flipped views:
$\bar{s}_i = \frac{s_i^{(o)} + s_i^{(f)}}{2}$,
where $s_i^{(o)}$ and $s_i^{(f)}$ denote the corresponding
confidence scores. A pseudo-label is accepted if its mean
confidence and geometric consistency satisfy the following
criteria:
\begin{equation}
a_i = \mathbb{1}\left[
\begin{aligned}
&\bar{s}_i \geq \tau_{\mathrm{conf}}
\;\land\;
C_i^{\mathrm{IoU}} \geq \tau_{\mathrm{IoU}}\\
&\land\;
\sigma_i^{\mathrm{center}} \leq \tau_{\mathrm{center}}
\end{aligned}
\right],
\label{eq:CCGP_acceptance}
\end{equation}
where $\mathbb{1}[\cdot]$ is the indicator function,
$C_i^{\mathrm{IoU}}$ measures bounding-box consistency, and
$\sigma_i^{\mathrm{center}}$ measures normalized center
disagreement across views. The thresholds are
$\tau_{\mathrm{conf}}=0.80$, $\tau_{\mathrm{IoU}}=0.75$,
and $\tau_{\mathrm{center}}=0.04$.

The final bounding box is obtained by confidence-weighted fusion of the original-view and restored flipped-view predictions:
\begin{equation}
\bar{b}_i =
\frac{
s_i^{(o)}\hat{b}_i^{(o)}
+s_i^{(f)}\tilde{b}_i^{(f)}
}{
s_i^{(o)}+s_i^{(f)}
}.
\label{eq:box_fusion}
\end{equation}
where, \(\bar{b}_i\) denotes the fused bounding box, while \(\hat{b}_i^{(o)}\) and \(\tilde{b}_i^{(f)}\) represent the original-view and restored flip-view predictions, respectively.
Among the 15,000 CelebA~\cite{liu2015faceattributes} candidates, 13,375 (89.17\%) were accepted, 1,462 rejected, and
163 yielded no detection. All rejected candidates failed the mean confidence threshold of 0.80. Although 107 failed the consistency IoU threshold, 49 exceeded the center disagreement threshold, and 90 had fewer than two valid views, none of these criteria rejected any additional confidence passing candidate. Thus, geometric consistency served as a safeguard rather than an active filter. The minimum per view confidence threshold of 0.25 was never violated.
Geometric checks did not reject additional samples; confidence-weighted fusion combined predictions from both views to generate the final pseudo-labels.
The accepted pseudo-label set is
\begin{equation}
\mathcal{D}_{p}
=
\left\{
(x_i,\hat{b}_i)
\mid x_i\in\mathcal{D}_{u},\;a_i=1
\right\},
\label{eq:pseudo_label_set}
\end{equation}
where $\hat{b}_i$ is the confidence-weighted fused bounding box.
Only accepted pseudo-labels are retained for subsequent training.
Cross-view consistency serves as an uncertainty proxy; it does
not guarantee correct localization.

\subsection{Detector Fine-Tuning}
\label{sec:detector_finetuning}

After pseudo-label selection, the accepted annotations are combined with the labeled data: 
$\mathcal{D}_{\mathrm{CCGP}} = \mathcal{D}_{l}\cup\mathcal{D}_{p}$.
We initialize a Grounding DINO model from the same supervised checkpoint used for pseudo-label generation and fine-tune it on $\mathcal{D}_{\mathrm{CCGP}}$. Let $\theta$ denote the parameters of the fine-tuned detector. Its training objective is the standard Grounding DINO detection objective: $\mathcal{L}_{\mathrm{CCGP}} = \mathcal{L}_{\mathrm{det}} \left(f_{\theta}, \mathcal{D}_{\mathrm{CCGP}} \right),$
where $\mathcal{L}_{\mathrm{det}}$ denotes the detection loss defined by the underlying Grounding DINO implementation.
CCGP does not introduce an additional detection loss or modify the detector architecture. Instead, it changes the composition of the training data by selecting pseudo labels according to detection confidence and cross view geometric consistency. The resulting finetuned detector can exploit additional training examples while limiting the
influence of pseudo labels that fail the selection criteria. This hypothesis is evaluated through controlled comparisons with supervised training and conventional confidence based self training.

%% file: sec/4_experiments.tex
\section{Experimental Setup}
\label{sec:experiments}
This section presents the datasets, evaluation protocols, and implementation details used to validate the proposed framework. The experiments are designed to evaluate forehead localization across diverse facial appearances
and imaging conditions. Multiple masked-face datasets are utilized to simulate realistic scenarios where the lower facial region is occluded, highlighting the importance of the forehead as a reliable biometric region for recognition.
\subsection{Datasets}
\label{sec:datasets}

\begin{figure}[htbp]
    \centering
    \includegraphics[width=0.5\textwidth]{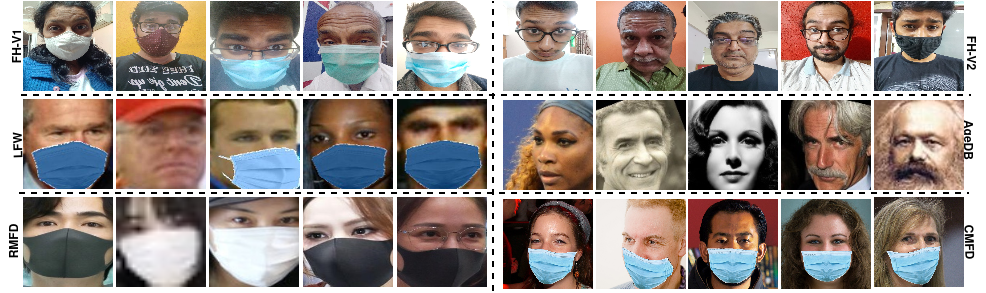}
    \caption{Real World Masked Face (RWMF) and Simulated Masked Face (RWSMF) datasets sample images}
    \label{fig:dataset_samples}
\end{figure}

We evaluate the proposed framework on multiple facial image datasets comprising masked and unmasked facial images as shown in Figure \ref{fig:dataset_samples}, Table \ref{tab:dataset_summary}. The datasets provide diverse facial appearances and imaging conditions to evaluate the localization of the forehead region. In particular, the inclusion of masked faces is relevant to our setting because facial masks occlude a substantial portion of the lower face, making the visible upper-face region, including the forehead, an important target 
for reliable localization.


\begin{table}[htbp]
\centering
\small
\caption{Datasets used for evaluation and pseudo-label generation. RWMF, RWSMF, and RWF denote real-world masked-face, simulated masked-face, and real-world face datasets.}
\label{tab:dataset_summary}
\resizebox{\columnwidth}{!}{
\begin{tabular}{lccccc}
\toprule
Dataset & Type & Subjects & Train & Val & Test \\
\midrule
FH-V1~\cite{bharadwaj2022mobile}
& RWMF & 247 & 2508 & 1250 & 1206 \\

CelebA~\cite{liu2015faceattributes}
& RWF & -- & 13,375 & -- & - \\

FH-V2~\cite{tandon2024synthetic}
& RWMF & 295 & -- & -- & 5890 \\

LFW~\cite{dalkran2021lfwmasked}
& RWSMF & 5749 & -- & -- & 13130 \\

AgeDB-30~\cite{moschoglou2017agedb}
& RWSMF & 568 & -- & -- & 16412 \\

RMFRD~\cite{wang2023masked}
& RWMF & 426 & -- & -- & 2095 \\

MaskedFace-Net~\cite{cabani2021maskedface}
& RWSMF & -- & -- & -- & 67047 \\
\bottomrule
\end{tabular}
}
\end{table}


\noindent\textbf{Forehead V1 (FH-V1) \cite{bharadwaj2022mobile}:}
The FH-V1\cite{bharadwaj2022mobile} dataset comprises 4,964 images collected from 247 subjects, including males, females, and children. The images capture masked and unmasked faces under varying facial expressions, providing diverse conditions for forehead analysis. The dataset supports the evaluation of forehead ROI localization and its suitability for subsequent biometric identification.

\noindent\textbf{CelebA}~\cite{liu2015faceattributes}:
CelebA contains 202,599 face images from 10,177 identities. We select 15,000 images for CCGP pseudo-label generation,
of which 13,375 accepted pseudo-labels are used to augment the manually annotated FH-V1 training data during fine-tuning.

\noindent\textbf{Forehead V2 (FH-V2) \cite{tandon2024synthetic}:}
FH-V2~\cite{tandon2024synthetic} comprises 5890 images forehead images from
295 subjects and is used exclusively for open-set evaluation.
Approximately 20 images per subject were acquired across two
sessions separated by two months using multiple mobile devices.
This dataset is used to evaluate the generalization of forehead
ROI localization to previously unseen subjects and across
variations in acquisition devices and sessions.

\noindent\textbf{LFW Simulated Masked Face (LFW-SMF)}
\cite{dalkran2021lfwmasked}:
LFW-SMF is derived from the Labeled Faces in the Wild (LFW)
benchmark, which contains 13,130 face images from 5,749 identities
collected under unconstrained, real-world conditions. In LFW-SMF,
synthetic masks are superimposed on the lower facial region of the
original images to simulate facial occlusion caused by mask wearing.
We use this benchmark to evaluate forehead ROI localization when
the lower facial region is occluded.

\noindent\textbf{AgeDB-30 Simulated Masked Face (AgeDB-30-SMF)}
\cite{moschoglou2017agedb}:
AgeDB-30 is a face verification benchmark containing 16,412 images
from 568 identities, with substantial variation in subjects' age,
pose, and appearance. We construct the simulated masked version
by introducing artificial masks over the lower facial region.
These images are used to assess the robustness of forehead ROI
localization under variations in image quality, pose, and other
challenging imaging conditions.

\noindent\textbf{MaskedFace-Net -- Correctly Masked Face Dataset (CMFD)}
\cite{cabani2021maskedface}: MaskedFace-Net is a synthetically generated masked-face dataset
derived from Flickr-Faces-HQ (FFHQ). It contains approximately 67,047 masked-face images with automatically generated masks that vary in shape, color, and placement. We use this dataset to evaluate the generalization of forehead ROI localization under diverse mask configurations and appearances.

\noindent\textbf{Real-World Masked Face Recognition Dataset (RMFRD)}
\cite{wang2023masked}: RMFRD contains real-world masked-face images representing 2095 images from 426 identities. Unlike synthetically masked benchmarks, it captures naturally occurring variations in mask appearance, facial pose, illumination, and image acquisition conditions. We use this dataset to evaluate the generalization of forehead ROI localization under naturally occurring facial occlusion.

The evaluation assesses forehead ROI localization accuracy and cross-dataset robustness, but not crease preservation or identity discrimination, which require separate segmentation and biometric recognition experiments.

\subsection{Evaluation Protocols}
\label{sec:evaluation_protocol}

This section describes the evaluation protocols used to assess localization accuracy, cross-dataset generalization, and robustness under varying image quality conditions.
\subsubsection{Subject-Disjoint Evaluation}
\label{sec:subject_disjoint}

We adopt a subject-disjoint split of FH-V1 into training, validation, and test subsets to prevent identity overlap.
The validation set is used for model selection, while the test set is reserved for final evaluation. For CCGP, the supervised Grounding DINO model is initialized using the 2,508 labeled FH-V1 training images. We additionally
select 15,000 unlabeled CelebA images for pseudo-label generation, retaining 13,375 images after confidence and consistency guided pseudo-label selection. The resulting training set contains 15,883 images, combining FH-V1 ground-truth annotations with accepted CelebA
pseudo-labels. FH-V1 validation and test images are excluded from pseudo-label generation and training. All methods are evaluated on the same FH-V1 test split using consistent evaluation settings to ensure a fair comparison.

\subsubsection{Cross-Dataset Evaluation}
\label{sec:cross_dataset}

To examine the generalization of forehead localization beyond the FH-V1 evaluation setting, we additionally evaluate the trained models on FH-V2, LFW, AgeDB, RMFRD, and CMFD, using the corresponding ground-truth annotations. These datasets are used exclusively for evaluation and are excluded from pseudo-label generation, model training, and checkpoint selection.
All models are evaluated using the same images, ground-truth annotations, detection confidence threshold, and evaluation metrics within each dataset. This protocol assesses the generalization of the proposed CCGP framework across datasets with different facial appearance characteristics and image acquisition conditions.

\subsubsection{Quality stratified robustness evaluation}
To evaluate the robustness of forehead ROI localization under diverse real world imaging conditions, we construct a quality stratified subset of 350 face images from five datasets: FH-V2\cite{tandon2024synthetic}, AgeDB \cite{moschoglou2017agedb}, LFW\cite{dalkran2021lfwmasked}, CMFD\cite{cabani2021maskedface}, and RMFRD\cite{wang2023masked}. The subset comprises seven categories, each containing $50$ images: blur, face scale variation, illumination variation, shadow, low contrast, off-angle pose, and good-quality images. Image selection was performed using a rule-based image quality assessment and face-pose estimation pipeline. Face detection and bounding-box measurements were obtained using a Haar Cascade face detector, while head-pose variations were estimated using 6DRepNet\cite{hempel20226d}, which provides pitch, yaw, and roll angles. Image quality was characterized using Laplacian variance for blur assessment, mean brightness for illumination assessment, intensity contrast for low-contrast detection, and a shadow score for shadow identification. The selection criteria included a blur score below $100$, a face-area ratio below $0.08$ or above $0.65$ for face scale variation, brightness below $55$ or above $205$ for illumination variation, contrast below $40$ for low contrast, and a shadow score above $55$ for shadow conditions. Images were categorized as off-angle when the absolute yaw exceeded $20°$, pitch exceeded $18°$, or roll exceeded $20°$. Images with detected faces that did not meet any of these criteria were assigned to the good quality category. 
All images were manually inspected to verify category suitability, facial visibility, and the presence of the target imaging condition, enabling systematic evaluation of forehead ROI localization under real-world image quality variations.
\noindent\textbf{Dataset Composition and Limitation:}
The quality stratified subset spans five datasets with varying domain characteristics. Since dataset composition may differ
across quality categories, observed performance differences may reflect both image quality and dataset specific effects.
Thus, the results assess robustness across imaging conditions rather than the isolated effect of image quality.

\subsection{Evaluation Metrics}
\label{subsec:evaluation_metrics}

We evaluate forehead ROI localization using mean Intersection
over Union (mIoU), precision, recall, and F1-score. These metrics
assess localization overlap and detection performance from
complementary perspectives.

\noindent\textbf{Intersection over Union (IoU):}
IoU measures the overlap between a predicted bounding box
$b_{\mathrm{pred}}$ and its ground-truth bounding box
$b_{\mathrm{gt}}$:
\begin{equation}
\operatorname{IoU}(b_{\mathrm{pred}},b_{\mathrm{gt}})
=
\frac{|b_{\mathrm{pred}}\cap b_{\mathrm{gt}}|}
{|b_{\mathrm{pred}}\cup b_{\mathrm{gt}}|}.
\label{eq:iou}
\end{equation}
Mean IoU (mIoU) is calculated by averaging the IoU values
across the evaluated images, without applying an IoU threshold
to the overlap calculation.

\noindent\textbf{Precision and Recall:}
Precision and recall quantify detection-level performance at
an IoU threshold of $\tau=0.75$. A predicted bounding box is
considered a true positive (TP) when it matches a ground-truth
box with IoU $\geq 0.75$. Unmatched predictions are counted
as false positives (FP), while unmatched ground-truth boxes
are counted as false negatives (FN). The metrics are defined as
\begin{equation}
\operatorname{Precision}
=
\frac{\mathrm{TP}}{\mathrm{TP}+\mathrm{FP}},
\label{eq:precision}
\end{equation}
\begin{equation}
\operatorname{Recall}
=
\frac{\mathrm{TP}}{\mathrm{TP}+\mathrm{FN}}.
\label{eq:recall}
\end{equation}

\noindent\textbf{F1-score:}
F1-score is the harmonic mean of precision and recall,
providing a single measure of their balance:
\begin{equation}
\operatorname{F1}
=
\frac{2\,\operatorname{Precision}\,
\operatorname{Recall}}
{\operatorname{Precision}+\operatorname{Recall}}.
\label{eq:f1}
\end{equation}
Unlike mIoU, precision, recall, and F1-score depend on the
detection matching threshold. We report 95\% confidence
intervals(CIs) for mIoU, while precision, recall, and F1-score
are reported as point estimates.

\section{Results \& Discussion}
\label{sec:results_discusiion}

 We evaluate CCGP in terms of localization accuracy, cross-dataset generalization, robustness, and computational efficiency.
\subsection{Subject-Disjoint Generalization}
\label{subsec:cross_dataset_generalization}

The proposed CCGP approach maintains competitive performance on the FH-V1 subject-disjoint test set, as shown in
Table~\ref{tab:SD_results}. While DiffusionInst achieves the highest mIoU (0.901), CCGP achieves 0.861 mIoU and an F1-score
of 0.946, closely approaching the supervised Grounding DINO baseline (0.866 mIoU and 0.948 F1-score). CCGP also achieves
a precision of 0.946 and a recall of 0.947, indicating balanced localization performance. However, the marginal difference in
mIoU suggests that pseudo-label filtering does not improve spatial overlap over the supervised baseline in this evaluation.
These findings indicate that CCGP preserves competitive localization performance, although its benefits are not reflected
in improved spatial overlap on the subject-disjoint test set.

\begin{table}[ht]
\centering
\caption{Forehead ROI localization performance on the FH-V1 subject-disjoint test set. mIoU is reported with 95\% confidence
intervals (CIs). Precision, recall, and F1-score are computed using detection matching at an IoU threshold of 0.75.}
\label{tab:SD_results}
\resizebox{\columnwidth}{!}{
\begin{tabular}{lcccc}
\toprule
\textbf{Method} & \textbf{mIoU} & \textbf{F1-score} & \textbf{Precision} & \textbf{Recall} \\
\midrule
YOLOv7\cite{wang2023yolov7} & 0.844[0.836,0.852] & 0.923 & 0.931 & 0.915 \\
YOLOv8\cite{jocher2023ultralytics} & 0.846[0.842,0.848] & 0.921 & 0.921 & 0.921 \\
Mask R-CNN\cite{he2017mask} & 0.855[0.851,0.858] & 0.936 & 0.936 & 0.936 \\
DiffusionInst\cite{gu2024diffusioninst} & \textbf{0.901}[0.896,0.906] & 0.938 & 0.937 & 0.938 \\
Grounding DINO\cite{liu2024eccv-grounding} & 0.866[0.863,0.870] & \textbf{0.948} & \textbf{0.948} & 0.947 \\
\textbf{Proposed (Ours)} & 0.861[0.858,0.865] & 0.946 & 0.946 & \textbf{0.947} \\
\bottomrule
\end{tabular}
}
\end{table}

\subsection{Cross-Dataset Generalization}
\label{subsec:cross_dataset_generalization}
Evaluating models trained on FH-V1 without target-dataset fine-tuning reveals substantial differences in cross-dataset generalization. As shown in Table~\ref{tab:cross_dataset}, the proposed CCGP detector achieves the highest mIoU on four of five external datasets: LFW (0.763), AgeDB (0.759), CMFD (0.722), and RMFRD (0.580). On FH-V2, it remains competitive with an mIoU of 0.709, although YOLOv8 performs slightly better (0.713). 
The gains are particularly pronounced on LFW, AgeDB, and RMFRD, where mIoU improves over the supervised Grounding DINO baseline by 0.108, 0.085, and 0.165, respectively. On RMFRD, the F1-score increases from 0.292 to 0.385, while on CMFD, it improves from 0.539 to 0.550. These results suggest that CCGP fine-tuning improves forehead localization across datasets with varying image characteristics. 
The precision and recall results further support these findings. On RMFRD, the proposed method achieves the highest recall (0.367) and improves precision from 0.378 to 0.405 compared with Grounding DINO. On LFW, it also improves F1-score (0.554 to 0.684), precision (0.592 to 0.688), and recall (0.521 to 0.681). CCGP improves cross-dataset localization, particularly
on challenging external datasets, although it does not outperform all baselines across every dataset and metric.

\begin{table}[htbp]
\centering
\caption{Cross-dataset forehead ROI localization performance. mIoU is reported with 95\% CIs. Precision,
recall, and F1-score are computed using detection matching at an IoU threshold of 0.75.}
\label{tab:cross_dataset}
\resizebox{\columnwidth}{!}{
\begin{tabular}{llcccc}
\toprule
\textbf{Method} & \textbf{Test Dataset} & \textbf{mIoU} & \textbf{F1-score} & \textbf{Precision} & \textbf{Recall} \\
\midrule
YOLOv7\cite{wang2023yolov7} & FH-V2 & 0.702[0.698,0.705] & 0.425 & 0.428 & 0.423 \\
YOLOv8\cite{jocher2023ultralytics} & FH-V2 & 0.713[0.709,0.716] & 0.467 & 0.468 & 0.467 \\
Mask R-CNN\cite{he2017mask} & FH-V2 & 0.700[0.697,0.704] & 0.424 & 0.425 & 0.423 \\
DiffusionInst\cite{gu2024diffusioninst} & FH-V2 & 0.701[0.697,0.704] & 0.417 & 0.418 & 0.417 \\
Grounding DINO\cite{liu2024eccv-grounding} & FH-V2 & 0.702[0.699,0.706] & 0.422 & 0.422 & 0.422 \\
\textbf{Proposed (Ours)} & \textbf{FH-V2} & \textbf{0.709[0.705,0.712]} & \textbf{0.451} & \textbf{0.452} & \textbf{0.451} \\
\midrule
YOLOv7\cite{wang2023yolov7} & LFW & 0.571[0.566,0.576] & 0.356 & 0.390 & 0.327 \\
YOLOv8\cite{jocher2023ultralytics} & LFW & 0.548[0.542,0.554] & 0.497 & 0.583 & 0.433 \\
Mask R-CNN\cite{he2017mask} & LFW & 0.647[0.644,0.650] & 0.257 & 0.265 & 0.249 \\
DiffusionInst\cite{gu2024diffusioninst} & LFW & 0.682[0.678,0.686] & 0.550 & 0.577 & 0.526 \\
Grounding DINO\cite{liu2024eccv-grounding} & LFW & 0.655[0.650,0.659] & 0.554 & 0.592 & 0.521 \\
\textbf{Proposed} & \textbf{LFW} & \textbf{0.763[0.760,0.766]} & \textbf{0.684} & \textbf{0.688} & \textbf{0.681} \\
\midrule
YOLOv7\cite{wang2023yolov7} & AgeDB & 0.433[0.430,0.440] & 0.186 & 0.217 & 0.162 \\
YOLOv8\cite{jocher2023ultralytics} & AgeDB & 0.517[0.512,0.522] & 0.310 & 0.367 & 0.269 \\
Mask R-CNN\cite{he2017mask} & AgeDB & 0.635[0.632,0.638] & 0.334 & 0.352 & 0.318 \\
DiffusionInst\cite{gu2024diffusioninst} & AgeDB & 0.537[0.532,0.542] & 0.403 & 0.475 & 0.350 \\
Grounding DINO\cite{liu2024eccv-grounding} & AgeDB & 0.674[0.670,0.677] & 0.514 & 0.538 & 0.492 \\
\textbf{Proposed} & \textbf{AgeDB} & \textbf{0.759[0.756,0.761]} & \textbf{0.660} & \textbf{0.662} & \textbf{0.659} \\
\midrule
YOLOv7\cite{wang2023yolov7} & CMFD & 0.485[0.482,0.488] & 0.413 & 0.506 & 0.349 \\
YOLOv8\cite{jocher2023ultralytics} & CMFD & 0.515[0.512,0.518] & 0.422 & 0.503 & 0.364 \\
Mask R-CNN\cite{he2017mask} & CMFD & 0.644[0.642,0.646] & 0.452 & 0.478 & 0.429 \\
DiffusionInst\cite{gu2024diffusioninst} & CMFD & 0.535[0.532,0.537] & 0.404 & 0.474 & 0.352 \\
Grounding DINO\cite{liu2024eccv-grounding} & CMFD & 0.699[0.698,0.701] & 0.539 & 0.554 & 0.525 \\
\textbf{Proposed} & \textbf{CMFD} & \textbf{0.722[0.720,0.723]} & \textbf{0.550} & \textbf{0.554} & \textbf{0.546} \\
\midrule
YOLOv7\cite{wang2023yolov7} & RMFRD & 0.330[0.315,0.346] & 0.225 & 0.307 & 0.178 \\
YOLOv8\cite{jocher2023ultralytics} & RMFRD & 0.388[0.372,0.404] & 0.320 & 0.432 & 0.254 \\
Mask R-CNN\cite{he2017mask} & RMFRD & 0.391[0.378,0.405] & 0.112 & 0.137 & 0.095 \\
DiffusionInst\cite{gu2024diffusioninst} & RMFRD & 0.334[0.318,0.350] & 0.293 & \textbf{0.451} & 0.217 \\
Grounding DINO\cite{liu2024eccv-grounding} & RMFRD & 0.415[0.400,0.430] & 0.292 & 0.378 & 0.238 \\
\textbf{Proposed} & \textbf{RMFRD} & \textbf{0.580[0.567,0.592]} & \textbf{0.385} & 0.405 & \textbf{0.367} \\
\bottomrule
\end{tabular}
}
\end{table}

\begin{figure}[htbp]
    \centering
    \includegraphics[width=0.25\textwidth]{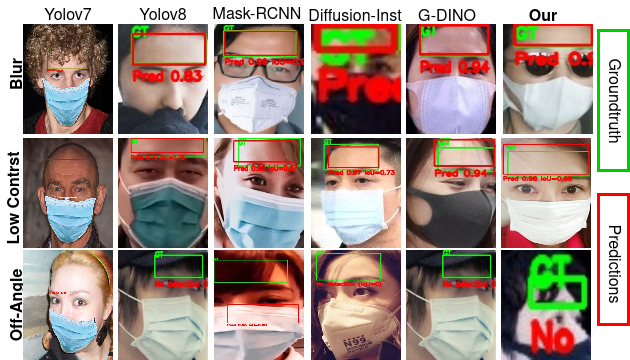}
\caption{Qualitative comparison of forehead ROI localization across blur, low contrast, and off-angle cases.}
\label{fig:qualitative_results}    
\end{figure}

\noindent\textbf{Qualitative Evaluation:}
Figure~\ref{fig:qualitative_results} compares forehead ROI localization under blur, low contrast, and off-angle pose. CCGP achieves competitive localization, although failures persist when forehead visibility is severely limited.

\noindent\textbf{Precision and Recall Analysis Across IoU Thresholds:}
The precision recall curves as shown in supplementary material reveal dataset dependent localization robustness.
CCGP achieves consistently stronger recall on LFW and AgeDB and the highest recall on RMFRD across most IoU thresholds, while maintaining competitive precision. On CMFD, its recall remains competitive with Grounding DINO, whereas FH-V2 shows closely overlapping curves across methods. Precision and recall decline sharply at stricter IoU thresholds, with all methods approaching zero at $\mathrm{IoU}\geq 0.95$. CCGP improves localization robustness on challenging external datasets, although its gains vary across datasets and IoU thresholds.
\subsection{Quality Stratified Robustness Under Real World Condition}
\label{subsec:quality_stratified_robustness}

Table~\ref{tab:robustness} summarizes localization performance across seven imaging
conditions. Compared with supervised Grounding DINO, CCGP improves mIoU in every evaluated stratum. The largest improvement occurs under blur, where mIoU increases from 0.487 to 0.719, followed by low contrast, where it increases from 0.615 to 0.710. Improvements are smaller under scale variation, illumination variation, and shadows. For good quality images, mIoU increases from 0.729 to 0.765. Off-angle pose remains challenging, with mIoU increasing from 0.458 to 0.555 but remaining the lowest among the seven conditions for CCGP. 
CCGP achieves higher mIoU than supervised Grounding DINO across all seven imaging conditions, with the largest gains under blur and low contrast. However, dataset-specific differences may also influence these results.


\begin{table}[ht]
\centering
\small
\caption{Forehead ROI localization robustness under real-world imaging conditions. mIoU is reported with 95\% CI.}
\label{tab:robustness}
\resizebox{\columnwidth}{!}{
\begin{tabular}{lcccccc}
\toprule
\textbf{Condition} & \textbf{YOLOv7} & \textbf{YOLOv8} & \textbf{Mask R-CNN} & \textbf{DiffusionInst} & \textbf{Grounding DINO} & \textbf{Our} \\
\midrule
Normal/Good & 0.611 & 0.650 & 0.687 & 0.608 & 0.729 & \textbf{0.765[0.727,0.797]} \\
Scale Variation & 0.412 & 0.416 & 0.584 & 0.464 & 0.628 & \textbf{0.663[0.606,0.712]} \\
Blur & 0.373 & 0.426 & 0.579 & 0.340 & 0.487& \textbf{0.719[0.671,0.762]} \\
Illumination & 0.566 & 0.575 & 0.685 & 0.599 & 0.727 & \textbf{0.759[0.721,0.792]} \\
Shadow & 0.384 & 0.474 & 0.536 & 0.452 & 0.615& \textbf{0.649[0.578,0.712]} \\
Low Contrast & 0.477 & 0.444 & 0.640 & 0.529 & 0.615 & \textbf{0.710[0.660,0.758]} \\
Off-Angle Pose & 0.260 & 0.324 & 0.468 & 0.339 & 0.458 & \textbf{0.555[0.491,0.620]} \\
\bottomrule
\end{tabular}
}
\end{table}

\subsection{Ablation Study}
\label{subsec:ablation_study}

Table~\ref{tab:rmfd_ablation} compares supervised Grounding DINO, conventional self-training, and CCGP on RMFRD (2,095 images).
Conventional self-training improves mIoU and F1-score from 0.415/0.292 to 0.551/0.382. CCGP further increases them to
0.580/0.385, with recall rising from 0.351 to 0.367, while precision decreases slightly from 0.419 to 0.405. These results
indicate that confidence \& consistency guided pseudo-label selection improves localization under cross-dataset shift, with a modest
precision recall trade-off. 
Conventional self-training selects single view Grounding DINO predictions from the same 30{,}000 image CelebA pool using a confidence threshold of $\tau=0.80$, yielding 24{,}518 pseudo-labels, whereas CCGP additionally incorporates cross-view consistency; both use identical fine-tuning settings.

\begin{table}[htbp]
\centering
\caption{Ablation study of forehead ROI localization on the RMFRD dataset.}
\label{tab:rmfd_ablation}
\resizebox{\columnwidth}{!}{
\begin{tabular}{lcccc}
\toprule
\textbf{Method} & \textbf{mIoU} & \textbf{F1-score} &
\textbf{Precision} & \textbf{Recall} \\
\midrule
Grounding DINO\cite{liu2024eccv-grounding}
& 0.415[0.400, 0.430] & 0.292 & 0.378 & 0.238 \\
Self-trained Grounding DINO
& 0.551[0.538, 0.564] & 0.382 & 0.419 & 0.351 \\
\textbf{CCGP (Ours)}
& \textbf{0.580[0.567, 0.592]} & \textbf{0.385}
& 0.405 & \textbf{0.367} \\
\bottomrule
\end{tabular}
}
\end{table}
\subsection{Efficiency Comparison}
\label{subsec:efficiency}
Table~\ref{tab:efficiency} compares the computational cost and inference throughput of the evaluated models. YOLOv7 and YOLOv8 achieve the highest throughput at 155.56 and 141.77 FPS, respectively, whereas Mask R-CNN and DiffusionInst incur higher computational costs and lower throughput. CCGP retains the Grounding DINO architecture and adds pseudo-label selection and fine-tuning. CCGP was evaluated on an NVIDIA H200 GPU with batch size 8, while the other models used batch size 1. These differences should be considered when interpreting FPS.
\begin{table}[ht]
\centering
\scriptsize
\setlength{\tabcolsep}{3pt}
\caption{Computational characteristics of the evaluated models.}
\label{tab:efficiency}
\begin{tabular}{@{}lccc@{}}
\toprule
\textbf{Method} & \textbf{Params (M)} & \textbf{FLOPs (G)} & \textbf{FPS} \\
\midrule
YOLOv7~\cite{wang2023yolov7} & 37.84 & 117.02 & 155.56 \\
YOLOv8~\cite{jocher2023ultralytics} & 11.78 & 42.44 & 141.77 \\
Mask R-CNN~\cite{he2017mask} & 43.92 & 372.96 & 31.45 \\
DiffusionInst~\cite{gu2024diffusioninst} & 112.26 & 369.61 & 19.22 \\
Grounding DINO~\cite{liu2024eccv-grounding} & 172.91 & 679.81 & 0.62 \\
\textbf{Ours} & 172.91 & 679.81 & 0.62 \\
\bottomrule
\end{tabular}
\end{table}
\noindent\textbf{Limitations.}
CCGP slightly underperforms supervised Grounding DINO in domain. The effects of confidence filtering, geometric consistency, and box fusion remain unisolated. Comparisons with advanced semi-supervised methods and independent pseudo-label quality assessment are also lacking. Robustness evaluation is limited by sample size and dataset composition. Moreover, geometric checks did not further reduce the accepted pseudo-label set at the selected thresholds.

\section{Conclusion}
\label{sec:conclusion}

We presented CCGP, a confidence and consistency guided pseudo-labeling framework for forehead localization using Grounding DINO. By combining detection confidence with cross view geometric consistency, CCGP filters pseudo-labels before fine-tuning on a combination of manually annotated and automatically labeled images. Experiments demonstrate improved cross-dataset localization on four of five external datasets, with the largest gains on RMFRD, where mIoU increase from 0.415 to 0.580 respectively. CCGP also improves localization under challenging imaging conditions, particularly blur and low contrast. 
Despite these gains, performance remains below the supervised baseline in-domain, and low inference throughput limits deployment. 